%% file: main.tex
\documentclass[10pt,twocolumn,letterpaper]{article}

 \usepackage[pagenumbers]{cvpr} 

\input{preamble}
\definecolor{cvprblue}{rgb}{0.21,0.49,0.74}
\usepackage[pagebackref,breaklinks,colorlinks,allcolors=cvprblue]{hyperref}
\usepackage{multirow}
\usepackage{amsmath}
\usepackage{marvosym}

\def\paperID{15572} 
\def\confName{CVPR}
\def\confYear{2026}

\title{UniTalking: A Unified Audio-Video Framework for Talking Portrait Generation}

\author{\parbox{18cm}{\centering
{\large Hebeizi Li $^{1,2}$\thanks{Equal contribution} $ $ \thanks{Work done as intern at Huawei.}, Zihao Liang $^{1*}$, Benyuan Sun$^{1 \ddag}$, Zihao Yin$^1$, Xiao Sha$^1$, Chenliang Wang$^1$, Yi Yang$^{1}$$^\textsuperscript{\Letter}$}\\
$^1$ Central Media Technology Institute, Huawei\\
$^2$ School of Computer Science and Engineering, Beihang University\\
$^\ddag$ Project Leader, $^\textsuperscript{\Letter}$ Corresponding Author}
}

\begin{document}
\maketitle


\input{secs/0_abstract}    
\input{secs/1_intro}
\input{secs/2_related}
\input{secs/data}
\input{secs/method}
\input{secs/experiments}
{
    \small
    \bibliographystyle{ieeenat_fullname}
    \bibliography{main}
}


\end{document}

%% file: secs/0_abstract.tex
\begin{abstract}

 While state-of-the-art audio-video generation models like Veo3 and Sora2 demonstrate remarkable capabilities, their closed-source nature makes their architectures and training paradigms inaccessible. To bridge this gap in accessibility and performance, we introduce UniTalking, a unified, end-to-end diffusion framework for generating high-fidelity speech and lip-synchronized video. At its core, our framework employs Multi-Modal Transformer Blocks to explicitly model the fine-grained temporal correspondence between audio and video latent tokens via a shared self-attention mechanism. By leveraging powerful priors from a pre-trained video generation model, our framework ensures state-of-the-art visual fidelity while enabling efficient training. Furthermore, UniTalking incorporates a personalized voice cloning capability, allowing the generation of speech in a target style from a brief audio reference. Qualitative and quantitative results demonstrate that our method produces highly realistic talking portraits, achieving superior performance over existing open-source approaches in lip-sync accuracy, audio naturalness, and overall perceptual quality.

\end{abstract}

%% file: secs/1_intro.tex
\section{Introduction}
\label{sec:intro}


The proliferation of Artificial Intelligence Generated Content (AIGC), largely propelled by advances in diffusion models, has demonstrated remarkable success in high-fidelity image, video, and audio synthesis. While initial research predominantly focused on single-modal generation, the frontier has recently shifted towards cross-modal tasks such as audio-to-video and video-to-audio synthesis. However, a critical and underexplored area is the unified generation of concurrently intertwined modalities. In real-world applications, video and audio are not merely sequential but synchronous, forming a cohesive perceptual experience. This underscores a significant gap in existing methodologies, which often fail to model this joint distribution effectively.

Existing approaches for joint audio-video generation reveal a significant divergence between large-scale proprietary models and accessible academic research. Recently, unreleased, closed-source systems such as Google's Veo3 and OpenAI's Sora2 have demonstrated state-of-the-art capabilities in generating highly coherent audio-visual content, including plausible talking portraits. However, the inaccessible nature of these models, which have undisclosed architectures, training data and methodologies, creates a critical barrier to academic progress and reproducibility. Consequently, the broader research community must still contend with existing open-source paradigms that suffer from fundamental limitations.

\begin{figure}[t]
    \centering
    \includegraphics[width=0.7\linewidth]{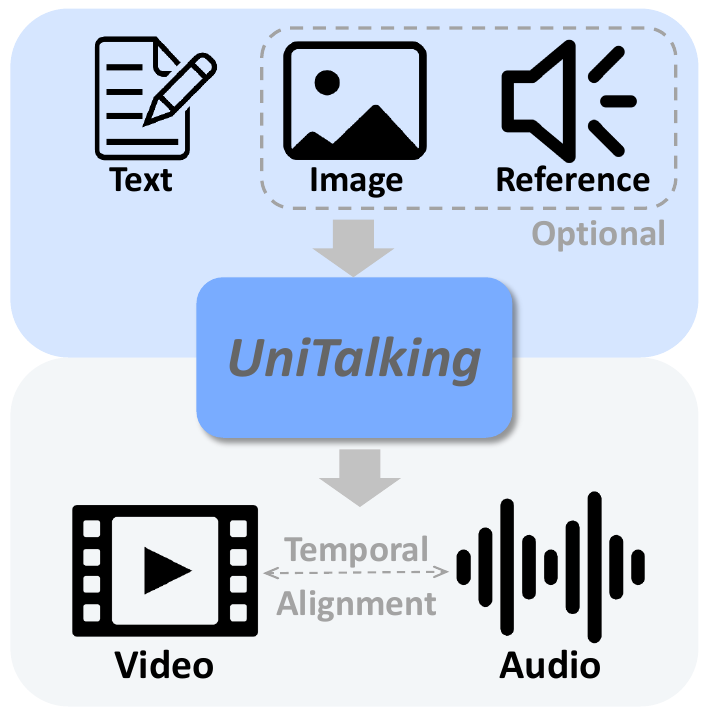}
    \caption{Illustration of our unified audio-video framework for talking portrait generation. UniTalking facilitates the generation of synchronized audio and video through multiple input modalities.}
    \label{fig:concept}
\end{figure}

Within the publicly accessible domain, these approaches are broadly categorized into two paradigms. The first employs a cascaded, two-stage process, where one modality (e.g., audio) is generated first and then used as a conditional signal to drive the synthesis of the other (e.g., video). While straightforward, this pipeline often struggles with temporal misalignment and error accumulation, leading to a noticeable lack of coherence. The second paradigm aims to directly generate both modalities simultaneously using an end-to-end model. However, a notable limitation of current work in this vein is its primary focus on synchronizing Foley sound with video, such as generating the visual of a crashing wave with its corresponding roar. These methods fall short of the precision required for complex tasks like speech. Therefore, developing an open and reproducible framework for generating talking portraits with frame-level, phonetically-precise lip synchronization remains a formidable and pressing challenge. This limitation severely curtails the applicability of AIGC in high-impact domains like automated film dubbing, realistic virtual avatars, and interactive digital humans.

To bridge this crucial gap, we introduce UniTalking, a novel framework for unified, end-to-end generation of talking portraits. As demonstrated in Figure \ref{fig:concept}, UniTalking facilitates the generation of unified audio-visual talking portraits and the production of personalized content based on reference images and audio.
The core of our framework is the Multi-Modal Transformer Block, which enforce fine-grained alignment between audio and visual modalities in the latent space. By employing shared self-attention mechanisms, this block explicitly models the temporal correspondence between visemes (visual) and phonemes (auditory), ensuring the generated lip movements are precisely synchronized with the synthesized speech. Furthermore, cross-attention with text conditions within the block guarantees that the generated content strictly adheres to the semantic meaning and lyrics information of the input text, achieving synchronized and semantically accurate generation.
In summary, our main contributions are as follows:
\begin{itemize}
    \item We introduce UniTalking, a novel end-to-end framework that unifies talking face video and corresponding speech audio generation within a symmetric dual-stream architecture, addressing a significant challenge in coherent audio-visual synthesis.
    \item We present a novel architectural design where a joint-attention mechanism within the Multi-Modal Transformer block operates on concatenated audio-visual tokens, enforcing fine-grained temporal alignment and achieving superior lip-sync accuracy.
    \item Our framework incorporates a multi-modal conditioning scheme, enabling the synthesis of talking portraits that preserve a specific visual identity from an image and mimic the vocal style of a reference audio clip.
    \item We establish a new state-of-the-art for open-source unified talking portrait generation, demonstrating superior performance in lip-sync, perceptual quality, and identity/style fidelity.
\end{itemize}

%% file: secs/2_related.tex
\section{Related Work}
\label{sec:related}
We situate our work within the context of recent advances in audio-visual generation. We categorize prior art into three primary areas: Audio-to-Video (A2V) synthesis, Video-to-Audio (V2A) synthesis, and the nascent field of unified audio-video generation. Many state-of-the-art methods in these areas leverage Latent Diffusion Models (LDMs)~\cite{rombach2022high} often with a Diffusion Transformer (DiT)~\cite{peebles2023scalable}, backbone and trained with objectives like Flow Matching~\cite{lipman2022flow}.

\subsection{Audio-to-Video Portrait Generation}
Audio-driven talking portrait generation has recently gained traction, largely propelled by advances in foundational video models~\cite{yang2024cogvideox, kong2024hunyuanvideo, wan2025wan}. Hallo3~\cite{cui2025hallo3} pioneers this direction by adapting CogVideoX~\cite{yang2024cogvideox} with dedicated attention mechanisms to enforce audio-visual synchronization and identity preservation. Similarly, Hunyuanvideo-Avatar~\cite{chen2025hunyuanvideo} builds upon HunyuanVideo~\cite{kong2024hunyuanvideo}, focusing on controllable expressions by computing cross-attention between audio features and facial regions. A significant body of work leverages the powerful Wan model~\cite{wan2025wan}. For instance, OmniHuman~\cite{lin2025omnihuman} enables control over expression intensity, while MultiTalk~\cite{kong2025let} synthesizes multi-person conversations by designing specific attention maps. These cascaded methods, while effective, generate video conditioned on pre-existing audio.

\subsection{Video-to-Audio Generation}
Video-to-Audio (V2A) generation provides a complementary approach, where a silent video is used to synthesize corresponding audio. This field is bifurcated into foley sound synthesis and speech generation.

For foley sound, early methods like Diff-Foley~\cite{luo2023diff} utilize contrastive pre-training to align audio-visual features. More recent models, such as FoleyCrafter~\cite{zhang2024foleycrafter}, MMAudio~\cite{cheng2025mmaudio}, and Kling-Foley~\cite{wang2025kling}, have adopted the MM-DiT framework, performing joint attention over multiple modalities to achieve state-of-the-art results. For video-to-speech (V2S), Faces2Voices~\cite{kim2025faces} employs a three-stage pipeline to model content, timbre, and prosody. DeepAudio~\cite{zhang2025deepaudio} uses a modular, multi-stage approach, while AudioGen-Omni~\cite{wang2025audiogen} extends the Kling-Foley~\cite{wang2025kling} architecture to incorporate lyrics, achieving high-quality speech synthesis.

\subsection{Unified Audio-Video Generation}

\begin{figure*}[ht!]
    \centering
    \includegraphics[width=1.0\textwidth]{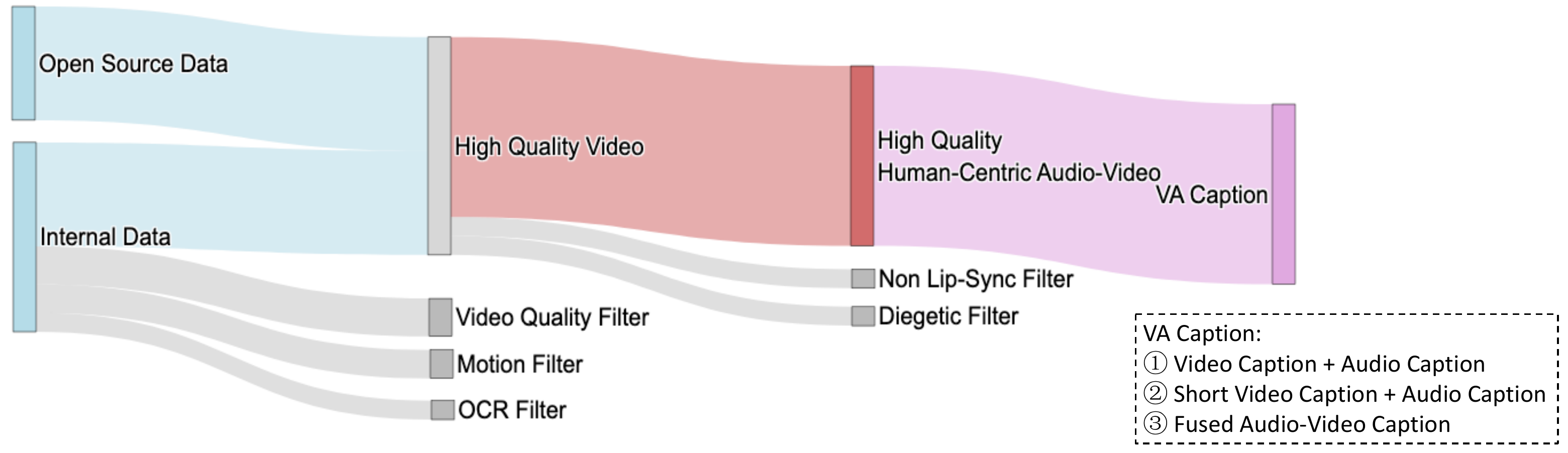}
    \caption{The data processing pipeline for curating our human-centric audio-video dataset. The pipeline employs sequential filtering, beginning with single-modality (video and audio) checks, followed by a cross-modal filtering stage. The final, high-quality audio-video pairs are then annotated with multi-level, multi-modal captions.}
    \label{fig:data}
\end{figure*}

Unified generation, which tackles both modalities simultaneously within a single model, is an emerging but critical research area. JavisDiT~\cite{liu2025javisdit} employs a dual-branch DiT with a Multi-Modality Bidirectional Cross-attention module for interaction. Universe-1~\cite{wang2025universe} leverages a Stitching of Experts (SoE) approach to integrate pre-trained unimodal models like Wan2.1~\cite{wan2025wan} and ACE-Step~\cite{gong2025ace}. Other works like Ovi~\cite{low2025ovi} and OmniTalker~\cite{wang2025omnitalker} also utilize dual-stream architectures, introducing dedicated fusion blocks or cross-attention layers for modal interaction.

UniTalking is distinct from these methods. While we also adopt a dual-stream architecture, our symmetric twin design, initialized from a powerful video foundation model, is specifically tailored for seamless modal alignment. Furthermore, unlike prior works that focus primarily on text-driven generation, our framework uniquely integrates multi-modal conditioning for both visual identity and acoustic style, enabling a new degree of personalization in a unified, end-to-end model.

%% file: secs/data.tex
\section{Data Preparation}

We construct our dataset from two primary sources: the open-source OpenHumanVid corpus \citep{li2025openhumanvid} and a large-scale, internal collection. This initial corpus undergoes a rigorous, multi-stage filtering pipeline designed to isolate high-quality, human-centric speech content (See Figure \ref{fig:data}). As detailed previously, this pipeline sequentially applies video, audio, and audio-video filtering modules to remove undesirable content (e.g., static video, low SNR audio, or poor lip-sync). Following curation, we generate multi-level textual annotations for each filtered audio-video pair. This annotation process produces captions at varying granularities. Except for captioning, we also generate reference audio for each video so that it can be used to train reference audio to audio-video generation. The final curated human-centric dataset consists of 2.3 million aligned audio-video samples.

\subsection{Data Processing Pipeline}

\noindent \textbf{Video-Audio Filtering. } We employ a sequential, three-stage filtering pipeline to curate the dataset:
\begin{itemize}
   \item Video Filtering: This initial pipeline processes the visual stream designed to filter out videos that are static, contain text overlays, or have low overall visual quality.
   \item Audio Filtering: The second pipeline analyzes the audio stream to remove samples that are muted, lack speech, or have a low signal-to-noise ratio (SNR). For this process, we utilize PANNs and the SentenceASD model to perform speech event detection.
   \item Audio-Video Filtering: The final pipeline assesses cross-modal relationships. It is applied to filter out videos containing purely diegetic audio (using LightASD) and samples exhibiting poor lip-synced alignment (using LipSync).
\end{itemize}

\noindent \textbf{Video-Audio Joint Captioning. } We generate three distinct textual captions to train our joint audio-video model. The first two formats are concatenated descriptions, where modality-specific captions are generated separately and then combined. These include: (a) a detailed video caption plus an audio caption, and (b) a short and concise video caption plus an audio caption. These independent descriptions are produced by a set of captioning models, including Qwen3-VL\citep{yang2025qwen3}, Whisper-V3\citep{radford2023robust}, and Qwen3-Omni-Captioner\citep{xu2025qwen3omni}. The third format is a fused description (c), for which we input both video and audio streams directly into Qwen3-Omni\citep{xu2025qwen3omni} to generate a single, unified audio-visual description. During training, we randomly sample one of the three prompts as input.

\subsection{Reference Data Generation}

In addition to standard Text-to-Video-Audio (T2VA) and Image-to-Video-Audio (TI2VA) generation, our model supports synthesis conditioned on a reference human voice to control the output timbre. As our source dataset lacks corresponding reference audio, we synthesize these samples to create the necessary training pairs.

We employ the IndexTTS2~\cite{zhou2025indextts2} model for this task. For each video, we generate a new audio clip by providing IndexTTS2 with two inputs: (1) the original audio from the video, which serves as the target timbre reference, and (2) a randomly generated text prompt. These prompts are constructed by randomly sampling sequences of words and numbers. We enforce a duration constraint, ensuring each synthesized reference audio clip is between 3 and 5 seconds. To augment our data, we repeat this process to generate three distinct reference audio samples for every video.

%% file: secs/method.tex
\begin{figure*}[ht]
    \centering
    \includegraphics[width=1.0\textwidth]{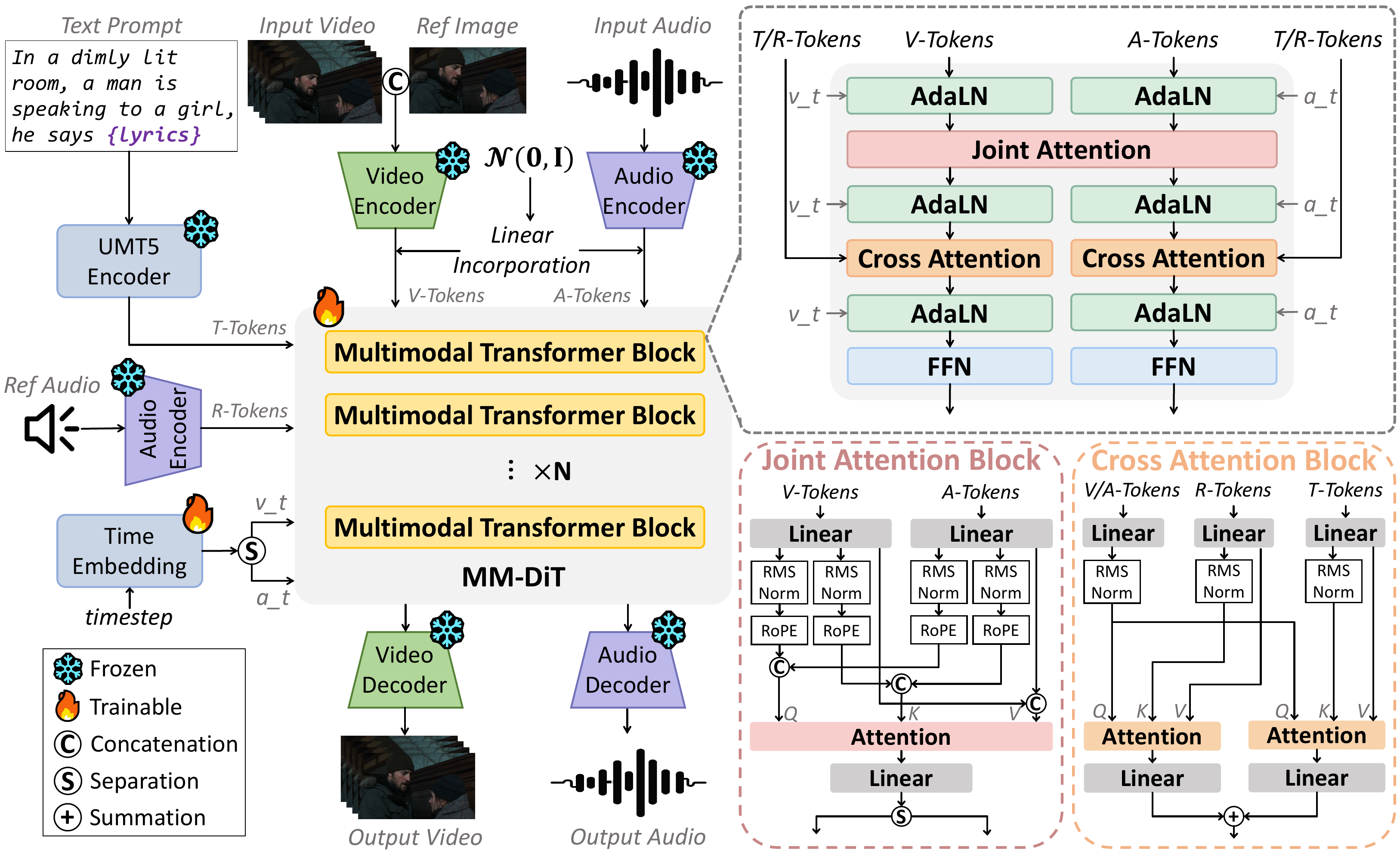}
    \caption{The architecture of UniTalking. Our framework jointly generates synchronized audio and video for talking portraits using a dual-stream Multi-Modal DiT (MM-DiT) backbone, which is trained as a Continuous Normalizing Flow via Flow Matching. The core of the model is the Multi-Modal Transformer Block, detailed on the right. It employs Joint Attention on concatenated audio-video tokens to ensure precise temporal alignment, and Cross Attention to incorporate multi-modal conditions such as text and acoustic style. Frozen components are marked with a snowflake, while trainable parts are marked with a flame.}
    \label{fig:main}
\end{figure*}

\section{Method}
In this section, we first introduce the foundational generative modeling principles our work is built upon. We then present the overall architecture of UniTalking, detail its core components for multi-modal processing, and finally, outline the progressive training strategy designed for efficient and stable convergence.

\subsection{Preliminaries}
\label{sec:preliminaries}
\noindent \textbf{Flow Matching.} Our framework is built upon Flow Matching~\cite{lipman2022flow}, a simulation-free method for training Continuous Normalizing Flows (CNFs). CNFs learn a time-dependent vector field $v_\theta (x_t, t)$ that defines a probability flow ODE, $\frac{\mathrm{d} x_t}{\mathrm{d} t} = v_\theta (x_t, t)$, which transforms a simple prior sample $x_0 \sim \mathcal{N} (0, \mathrm{I})$ at $t=0$ into a data sample $x_1$ at $t=1$. The model is trained directly using the Conditional Flow Matching (CFM) objective, which regresses onto the target vector $x_1-x_0$ along a simple linear path $x_t = (1-t)x_0+tx_1$:
\begin{equation}
    \mathcal{L}_{CFM} = \mathbb{E}_{t, q(x_1), p_0(x_0)} \left [ {\left \| v_\theta (x_t-t)-(x_1-x_0) \right \| }^2  \right ].
\end{equation}
Generation is performed by sampling $x_0$ and solving the learned ODE with a numerical solver.
In the proposed framework, the objective function for standalone audio generation tasks is defined as $L_{total} = L_{CFM}^a$, whereas for other joint generation tasks, the objective function is defined as $L_{total} = L_{CFM}^a + L_{CFM}^v$.

\noindent \textbf{Classifier-Free Guidance.} We employ Classifier-Free Guidance (CFG)~\cite{ho2022classifier} to control synthesis. The vector field $v_\theta(x_t, t, c)$ is trained on a condition $c$ which is randomly replaced by a null token $\emptyset $. At inference, the guided vector field $\hat{v_\theta}$ is extrapolated as:
\begin{equation}
    \hat{v_\theta}(x_t, t, c) = v_\theta(x_t, t, \emptyset) + \omega \cdot (v_\theta(x_t, t, c)-v_\theta(x_t, t, \emptyset)),
\end{equation}
where $\omega>1$ is the guidance scale. In UniTalking, condition $c$ can be constituted by a combination of text prompts ($c_{text}$), identity images ($c_{image}$) and reference audio ($c_{audio}$), thus enabling multiple tasks including text-to-audio-video (T2AV), text-image-to-audio-video (TI2AV), and text-reference-to-audio-video (TR2AV).

\subsection{Overall Architecture}
The architecture of UniTalking, illustrated in Figure~\ref{fig:main}, is based on the Multi-Modal Diffusion Transformer (MM-DiT)~\cite{esser2024scaling}. The entire framework is trained as a Continuous Normalizing Flow using the Flow Matching objective and steered by Classifier-Free Guidance, as detailed in Section~\ref{sec:preliminaries}. The core of our model is a transformer backbone comprising $N$ MM-DiT blocks, which operate on a concatenated sequence of latent video and audio tokens. These tokens are processed through a symmetric, dual-stream design within each block, a choice made to facilitate the learning of shared temporal dynamics. The transformer's output, a denoised latent representation, is subsequently decoded into coherent video frames and audio waveforms.

To leverage the power of large-scale pre-training, the video stream of our model inherits both the architecture and the weights of Wan2.2-5B~\cite{wan2025wan}, providing a powerful prior for visual synthesis. Critically, to bridge the representational gap between modalities, the audio stream is designed as an identical twin to the video stream. This architectural symmetry encourages seamless latent-space fusion. The parameters for this audio branch are randomly initialized and trained to match the capabilities of its pre-trained visual counterpart.

\subsection{Latent Representation} 
To ensure computational efficiency and focus learning on the core generative task, all input modalities are projected into a compact latent space by dedicated, pre-trained encoders that remain frozen throughout training.

\noindent \textbf{Video and Audio Latents.} Video frames are encoded by the 3D causal VAE~\cite{wu2025improved} from Wan2.2, achieving a high spatio-temporal compression of $16\times16\times4$. For audio, we employ the 1D VAE from MMAudio~\cite{cheng2025mmaudio}.
The raw waveform is converted to a Mel-spectrogram via Short-Time Fourier Transform (STFT)~\cite{stevens1937scale} and then encoded. During inference, the denoised audio latent is decoded back to a Mel-spectrogram and synthesized into a 44.1kHz waveform using the BigVGAN vocoder~\cite{lee2022bigvgan}.

\noindent \textbf{Conditioning Latents.} The model is guided by multi-modal conditions. Text prompts are embedded using a frozen UMT5 model~\cite{chung2023unimax}, while reference audio clips are encoded by the same frozen MMAudio VAE. For training stability, the sequence lengths for text and reference audio latents are fixed to 512 and 257, respectively. It is noted that longer sequences are truncated, while shorter ones are padded with zeros. This frozen-encoder setup provides a stable, semantically rich latent space for the MM-DiT blocks. 

\subsection{Multi-Modal Transformer Block} 
To effectively integrate our multi-modal conditions and enforce temporal coherence, we adapt the standard DiT block with threefold modifications targeting its attention mechanisms and positional encoding.

First, the standard self-attention is replaced with a joint-attention mechanism. Latent tokens from both video and audio streams are concatenated, forcing the model to learn intra- and inter-modal dependencies within a single attention operation to promote holistic audio-visual understanding.

Second, to jointly condition on text ($c_{text}$) and reference audio style ($c_{audio}$), we augment the cross-attention module. A supplementary set of key-value projection layers is introduced for the reference audio condition. Latent tokens from the main streams attend to each condition separately, and their outputs are fused via element-wise summation, enabling generation that is simultaneously faithful to the text and stylistically consistent with the reference audio.

Third, we employ a specialized Rotary Positional Embedding (RoPE)~\cite{su2024roformer} strategy. Following OVI~\cite{low2025ovi}, we apply standard RoPE along the temporal ($t$) axis. For the spatial ($h$, $w$) dimensions of audio tokens, however, we use a RoPE derived from a single, fixed position. This anisotropic design compels the model to prioritize temporal dynamics, fostering a stronger alignment between the audio and video streams. Further details are provided in the Appendix.

\subsection{Training Strategy}
Due to the initialization imbalance between the pre-trained video branch and the randomly initialized audio branch, we propose a progressive two-stage training strategy for stable convergence.

\noindent \textbf{Audio Branch Pre-training.} Our training pipeline begins by training the audio branch in isolation on a text-to-speech (TTS) task. We train it on a composite TTS task, jointly optimizing synthesis for both with and without timbre-referenced. To preserve the representations of other modalities, we adopt a parameter-efficient approach: all parameters related to the text and video branches (including their FFN and attention projection layers) are frozen. Only the audio input projection layers, audio branch FFN and attention projection layers are tuned. We find out that solely finetuning these parameters are sufficient to generate good speech voice. This stage compels the audio branch to learn a meaningful mapping from semantics to acoustics, aligning its representational capacity with the text branch. Our experiments find out that this stage is essential for the generated audio quality in the final audio-video generation.

\noindent \textbf{Multi-Task Joint Audio-Video Training.} With a capable audio branch, we train the entire UniTalking framework end-to-end. The objective here is to learn the intricate temporal correlations between audio and video. We employ a multi-task learning objective, alternating between tasks like Text-to-Audio-Video (T2AV), Text-Video-to-Audio (TV2A), Text-Image-to-Audio-Video (TI2AV) and Text-Reference-to-Audio-Video (TR2AV) to force the model to learn the bidirectional relationship between modalities, leading to precise lip synchronization. We posit that this alternation creates a powerful synergistic learning dynamic, forcing the model to learn a holistic and disentangled representation of audio-visual speech.

T2AV is the core joint generation task where the model synthesizes both audio and video from only a text prompt. It establishes a coarse-grained alignment, as the generated content must correspond to the shared underlying script. TV2A provides a strict, unidirectional supervisory signal for temporal alignment.
In this task, we incorporated an attention mask from audio tokens to video tokens, thereby preventing audio from influencing the video branch and enabling audio prediction based on the latent features of the actual video. This task compels the audio branch to learn the precise acoustic consequences of subtle, frame-by-frame visual movements, directly teaching the model the fine-grained mapping from visemes to phonemes. While TI2AV and TR2AV grant models the ability to generate personalized content.

By alternating between these four tasks, we create a multi-faceted set of constraints that prevent the model from finding simplistic solutions. This comprehensive strategy forces the model to learn the true underlying manifold of synchronized, controllable, and personalized audio-visual speech, leading to a robust and versatile final model.

%% file: secs/experiments.tex
\section{Experiments}
\subsection{Implementation Details}
In accordance with the multi-stage training strategy, the initial stage of audio branch pre-training is conducted with a batch size of 256 and a learning rate of $1\times 10^{-5}$ for a total of 100,000 steps on internal TTS data. The second stage of the joint video-audio training is conducted with a batch size of 64 and a learning rate of $1\times 10^{-5}$ for 100,000 steps. The training process is optimized through the utilization of the AdamW optimizer, with $\beta_1=0.9, \beta_2=0.999, \epsilon=10^{-8}$. The scheduling of training is based on the UniPC~\cite{zhao2023unipc} implementation. In order to achieve efficient training, it is imperative that all training utilizes $\mathtt{bf16}$ precision and is parallelized using Fully Sharded Data Parallel (FSDP)~\cite{zhao2023pytorch}. 

Given that the model is based on Wan2.2, there is a high degree of congruence between the framework details of the two structures. For instance, the number of MM-Dit Blocks is $N=30$, the model dimension is configured as $dim=3072$, and attention employs $24$ heads. Furthermore, the total number of parameters across the entire model is $10B$.

\begin{figure*}[t]
    \centering
    \includegraphics[width=1.0\linewidth]{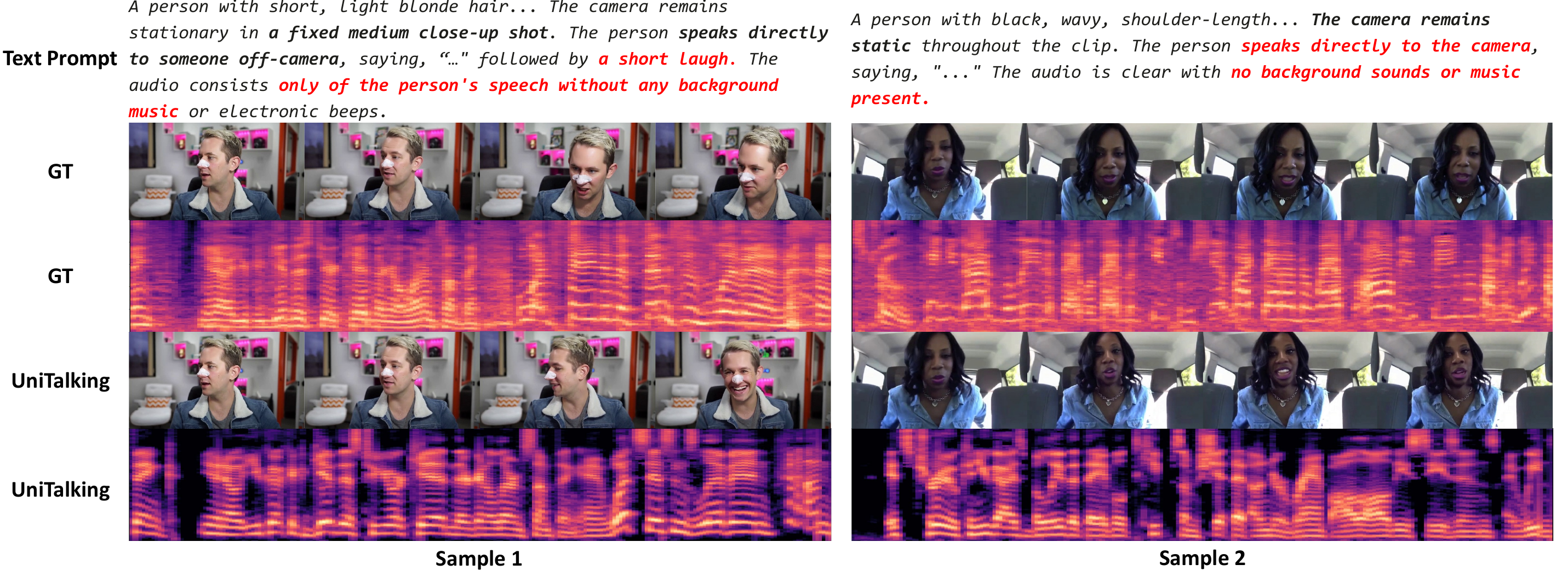}
    \caption{Visualization of video frames and Mel-spectrograms for generated audio-visual talking portraits and ground-truth audio-visual data. At the top is the corresponding text prompt, with certain keywords highlighted in \textbf{bold} and \textcolor{red}{\textbf{red}}. The generated video and audio are semantically consistent with the ground truth.}
    \label{fig:vis}
\end{figure*}

\subsection{Result Visualization}
\noindent \textbf{Audio-Video Synchronization.}
A fundamental challenge in the realm of unified audio-visual generation pertains to the temporal alignment between these two modalities. In order to demonstrate the alignment quality of the model's generation results, the video frames and audio Mel-spectrograms of randomly selected samples are visualized in Figure \ref{fig:vis}. By comparing the video frames with the Mel-spectrogram changes, it can be observed that the talking Portraits in the generated videos exhibit accurate lip-syncing. Furthermore, a comparison of the Mel-spectrograms of audio generated by UniTalking with those of ground truth reveals that both exhibit highly similar patterns, indicating that the generated audio is entirely accurate. The text prompt $\mathtt{"without\ any\  background \ music"}$ results in the generation of a video devoid of background noise, thereby enabling the Mel-spectrogram to accentuate vocal components with greater clarity. The prompt $\mathtt{"a\ short\ laugh"}$ in the text prompt also corresponds to the final video frame in Sample 1, demonstrating the model's capacity to adhere to the input text when generating results.

\begin{figure*}[t]
    \centering
    \includegraphics[width=0.8\linewidth]{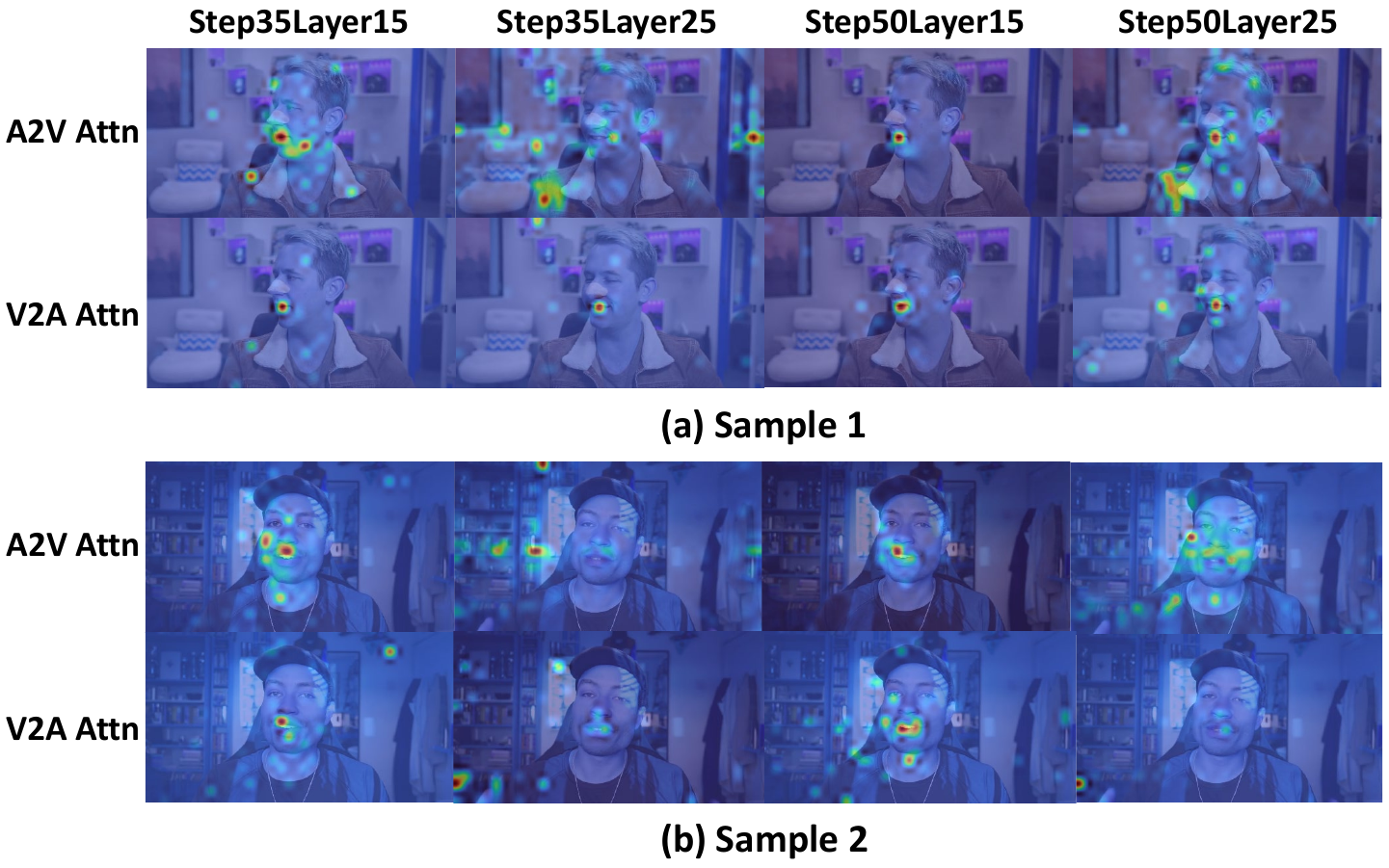}
    \caption{Visualization of attention maps between video tokens and audio tokens in Joint Attention. Subfigures (a) and (b) show two different random samples. The first row in the subgraph represents audio-to-video attention, while the second row represents video-to-audio attention. Different columns correspond to weights derived from distinct inference sampling steps and different transformer blocks.}
    \label{fig:attn}
\end{figure*}

\noindent \textbf{Cross-Attention Visualization.}
To further illustrate the correlation between video and audio modalities in UniTalking, attention maps cross video and audio tokens through Joint Attention module in different layers are presented in Figure \ref{fig:attn}. The attention maps reveal clear dependency between video and audio modalities, with higher weights assigned to semantically related regions.
The audio-to-video attention maps indicate that facial and body areas benefit from audio tokens. Conversely, the video-to-audio attention maps demonstrate that audio tokens benefit exclusively from the lip region. The correlation pattern is consistent across models with different training iterations. Additionally, some mismatched highlighted regions in attention maps change randomly with different training iterations. This is speculated to result from model defects caused by training strategies or data noise.

\subsection{Experiment Settings}
\noindent \textbf{T2AV Joint Generation.}
For audio-video joint generation, we compare with recent SOTA methods Universe-1~\cite{wang2025universe} and OVI~\cite{low2025ovi}. We conducted a blind pairwise preference study with 20 participants for each model pair comparison. The testset includes 50 test prompts for talking portraits generation. 

\noindent \textbf{TR2AV Joint Generation.}
For the TR2AV task, the most critical evaluation metric is the consistency between generated content and reference content. In this study, the experimental settings of Qwen3-Omni~\cite{xu2025qwen3omni} are employed, with the MiniMax Multilingual Test Set~\cite{zhang2025minimax} being utilized to evaluate the speaker similarity between the generated audio and the reference audio. 

\noindent \textbf{TTS Generation.}
For TTS generation, we compare with both single modality TTS model and OVI~\cite{low2025ovi}. We follow the evaluation protocol of TTS in OVI and test the model on Seed-TTS test-en dataset~\cite{anastassiou2024seed}. WER is used as evaluation metric.

\begin{figure}[h]
    \centering
    \includegraphics[width=0.95\linewidth]{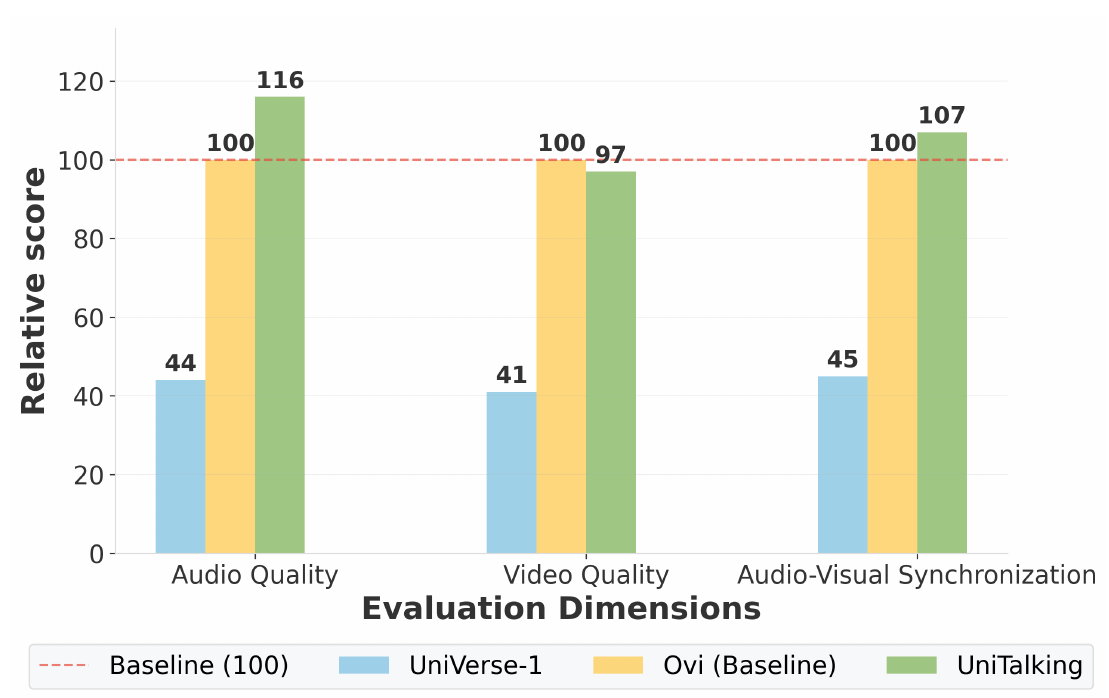}
    \caption{Results of the blind preference study for three metrics: video quality, audio quality, and audio-video synchronization. Ovi's results are normalized to 100 as the baseline, with other methods reporting relative scores.}
    \label{fig:rate}
\end{figure}

\vspace{-1.0em}
\subsection{Results}

\noindent \textbf{Blind Preference Study on T2AV.} 
We perform subjective evaluations on the T2AV generation, we use OVI~\cite{low2025ovi} as the baseline model and compute the win-lose rate for all model on video quality, audio quality, and audio-visual synchronization. The win-lose rate is computed as $rate = \frac{win + tie}{lose + tie}$. We construct a testset with 50 talking test prompts. As can be seen in Figure \ref{fig:rate}, our methods works best on the audio quality and audio-visual synchronization. Compared to OVI, the proposed method gets a 116\% and 107\% on audio Quality and audio-Visual synchronization. For the video quality, our model is on par with OVI. Since both of OVI and our method utilize the pretrained Wan 2.2 model, this performance is relatively reasonable.



We also provide an objective comparison using the Sync-C and Sync-D lip-sync metrics. As detailed in Table \ref{tab:lipsync}, our method surpasses Universe-1 by 3.02 points on Sync-C and achieves performance competitive with the Sora2 model. For Sync-D, we surparsses Universe-1 and OVI with 3.92 and 0.55. In this analysis, we observe that the OVI model obtains a notably high Sync-C score. We hypothesize this is due to a potential bias in the metric, as Sync-C appears to favor generated subjects with exaggerated mouth movements over more naturalistic articulation.

\begin{table}[ht]
\caption{Results of Lip-sync evaluation on T2AV}
\label{tab:lipsync}
\centering
\begin{tabular}{lcc}
\toprule
 Method & Sync-C $\uparrow$ & Sync-D $\downarrow$ \\ 
\midrule
Universe-1~\cite{wang2025universe} & 1.85 & 11.97 \\
OVI~\cite{low2025ovi} & 6.56 & 8.6 \\
Sora2 & 5.35 & 7.78 \\
\textsc{UniTalking} (Ours) & 4.87 & 8.05 \\
\bottomrule
\end{tabular}
\vspace{-1.5em}
\end{table}

\begin{table}[ht]
\centering
\caption{Results of Speaker Similarity on TR2AV. The text prompts from UniTalking supplement the video descriptions.}
\label{tab:tr2av}
\begin{tabular}{lll}
\toprule
\multirow{2}{*}{Method} & \multicolumn{2}{l}{Speaker Similarity $\uparrow$} \\ 
                        & English            & Chinese           \\ \midrule
ElevenLabs                 & 0.613              & 0.677             \\
MiniMax~\cite{zhang2025minimax}                 & 0.756              & 0.780             \\
Qwen3-Omni~\cite{xu2025qwen3omni}              & 0.773              & 0.772             \\
\textsc{UniTalking} (Ours)                    & 0.703              & 0.662             \\ \bottomrule
\end{tabular}
\end{table}

\noindent \textbf{Evaluation of Similarity to Reference.}
Table \ref{tab:tr2av} presents the results of our tonal similarity comparison experiment. Consistent with Qwen3-Omni~\cite{xu2025qwen3omni}, we selected ElevenLabs, MiniMax~\cite{zhang2025minimax}, and Qwen3-Omni~\cite{xu2025qwen3omni} as comparison methods. It can be seen that our approach achieves a certain degree of phonetic cloning capability. Although it is difficult to surpass the performance of the state-of-the-art large audio models, UniTalking achieves speech similarity comparable to ElevenLabs.

\noindent \textbf{Text-to-Speech Generation Evaluation.} 
Following the evaluation setting of OVI ~\cite{low2025ovi}, we assess the Text-to-Speech (TTS) generation capabilities of our stage-1 model. As shown in Table \ref{tab:tts}, our model achieves performance comparable to state-of-the-art (SOTA) methods, demonstrating the efficacy of our audio training pipeline in generating high-fidelity speech. Furthermore, an ablation study reveals that omitting this stage-1 training results in significant degradation of the audio generated by the final audio-video model. We hypothesize this is due to an imbalance in modal convergence rates: the video branch benefits from extensive pre-training on large-scale data, whereas the audio branch is comparatively under-trained.

\begin{table}[ht]
\caption{Results of TTS evaluation. We follow the evaluation protocol of OVI~\cite{low2025ovi}.}
\label{tab:tts}
\centering
\begin{tabular}{ccc}
\toprule
 Type & Method & WER $\downarrow$ \\ 
\midrule

\multirow{4}{*}{TTS} 
 & Fish Speech~\cite{liao2024fish}    & \textbf{0.008} \\
 & F5-TTS~\cite{chen2025f5}          & 0.018 \\
 & CosyVoice~\cite{du2024cosyvoice}         & 0.034 \\
 & FireRedTTS~\cite{guo2024fireredtts}        & 0.038 \\
\midrule
\multirow{2}{*}{Unified} 
 & Ovi-Aud~\cite{low2025ovi} & 0.035\\
 & \textsc{UniTalking} (Ours) & 0.038 \\
\bottomrule
\end{tabular}
\end{table}

\section{Conclusion}
In this work, we presented UniTalking, a unified framework for multi-task audio-video generation supporting T2VA, TI2VA, and TR2VA. We utilize a unified attention mechanism that directly aligns text, video, and audio modalities within a shared latent space. This approach departs from traditional twin-backbone networks that depend on cross-attention. Experimental results demonstrate the effectiveness of our method. Although validated on talking person generation, we posit that the UniTalking framework is broadly applicable to general audio-video synthesis, including sound effects and music.

\noindent \textbf{Future Work} The performance of our model is constrained by available training resources and data scale, particularly when compared to closed-source models. Furthermore, the framework does not yet support multi-person reference generation, such as the "Cameo" in Sora2, which leverages multiple audio-video clips as input. We remain these chanlleges as future work.

%% file: main.bbl
\begin{thebibliography}{42}
\providecommand{\natexlab}[1]{#1}
\providecommand{\url}[1]{\texttt{#1}}
\expandafter\ifx\csname urlstyle\endcsname\relax
  \providecommand{\doi}[1]{doi: #1}\else
  \providecommand{\doi}{doi: \begingroup \urlstyle{rm}\Url}\fi

\bibitem[Anastassiou et~al.(2024)Anastassiou, Chen, Chen, Chen, Chen, Chen, Cong, Deng, Ding, Gao, et~al.]{anastassiou2024seed}
Philip Anastassiou, Jiawei Chen, Jitong Chen, Yuanzhe Chen, Zhuo Chen, Ziyi Chen, Jian Cong, Lelai Deng, Chuang Ding, Lu Gao, et~al.
\newblock Seed-tts: A family of high-quality versatile speech generation models.
\newblock \emph{arXiv preprint arXiv:2406.02430}, 2024.

\bibitem[Chen et~al.(2025{\natexlab{a}})Chen, Liang, Zhou, Huang, Ma, Tang, Lin, Zhou, and Lu]{chen2025hunyuanvideo}
Yi Chen, Sen Liang, Zixiang Zhou, Ziyao Huang, Yifeng Ma, Junshu Tang, Qin Lin, Yuan Zhou, and Qinglin Lu.
\newblock Hunyuanvideo-avatar: High-fidelity audio-driven human animation for multiple characters.
\newblock \emph{arXiv preprint arXiv:2505.20156}, 2025{\natexlab{a}}.

\bibitem[Chen et~al.(2025{\natexlab{b}})Chen, Niu, Ma, Deng, Wang, JianZhao, Yu, and Chen]{chen2025f5}
Yushen Chen, Zhikang Niu, Ziyang Ma, Keqi Deng, Chunhui Wang, JianZhao JianZhao, Kai Yu, and Xie Chen.
\newblock F5-tts: A fairytaler that fakes fluent and faithful speech with flow matching.
\newblock In \emph{Proceedings of the 63rd Annual Meeting of the Association for Computational Linguistics (Volume 1: Long Papers)}, pages 6255--6271, 2025{\natexlab{b}}.

\bibitem[Cheng et~al.(2025)Cheng, Ishii, Hayakawa, Shibuya, Schwing, and Mitsufuji]{cheng2025mmaudio}
Ho~Kei Cheng, Masato Ishii, Akio Hayakawa, Takashi Shibuya, Alexander Schwing, and Yuki Mitsufuji.
\newblock Mmaudio: Taming multimodal joint training for high-quality video-to-audio synthesis.
\newblock In \emph{CVPR}, pages 28901--28911, 2025.

\bibitem[Chung et~al.(2023)Chung, Constant, Garcia, Roberts, Tay, Narang, and Firat]{chung2023unimax}
Hyung~Won Chung, Noah Constant, Xavier Garcia, Adam Roberts, Yi Tay, Sharan Narang, and Orhan Firat.
\newblock Unimax: Fairer and more effective language sampling for large-scale multilingual pretraining.
\newblock \emph{arXiv preprint arXiv:2304.09151}, 2023.

\bibitem[Cui et~al.(2025)Cui, Li, Zhan, Shang, Cheng, Ma, Mu, Zhou, Wang, and Zhu]{cui2025hallo3}
Jiahao Cui, Hui Li, Yun Zhan, Hanlin Shang, Kaihui Cheng, Yuqi Ma, Shan Mu, Hang Zhou, Jingdong Wang, and Siyu Zhu.
\newblock Hallo3: Highly dynamic and realistic portrait image animation with video diffusion transformer.
\newblock In \emph{CVPR}, pages 21086--21095, 2025.

\bibitem[Du et~al.(2024)Du, Chen, Zhang, Hu, Lu, Yang, Hu, Zheng, Gu, Ma, et~al.]{du2024cosyvoice}
Zhihao Du, Qian Chen, Shiliang Zhang, Kai Hu, Heng Lu, Yexin Yang, Hangrui Hu, Siqi Zheng, Yue Gu, Ziyang Ma, et~al.
\newblock Cosyvoice: A scalable multilingual zero-shot text-to-speech synthesizer based on supervised semantic tokens.
\newblock \emph{arXiv preprint arXiv:2407.05407}, 2024.

\bibitem[Esser et~al.(2024)Esser, Kulal, Blattmann, Entezari, M{\"u}ller, Saini, Levi, Lorenz, Sauer, Boesel, et~al.]{esser2024scaling}
Patrick Esser, Sumith Kulal, Andreas Blattmann, Rahim Entezari, Jonas M{\"u}ller, Harry Saini, Yam Levi, Dominik Lorenz, Axel Sauer, Frederic Boesel, et~al.
\newblock Scaling rectified flow transformers for high-resolution image synthesis.
\newblock In \emph{ICML}, 2024.

\bibitem[Gong et~al.(2025)Gong, Zhao, Wang, Xu, and Guo]{gong2025ace}
Junmin Gong, Sean Zhao, Sen Wang, Shengyuan Xu, and Joe Guo.
\newblock Ace-step: A step towards music generation foundation model.
\newblock \emph{arXiv preprint arXiv:2506.00045}, 2025.

\bibitem[Guo et~al.(2024)Guo, Hu, Liu, Shen, Tang, Wu, Xie, Xie, and Xu]{guo2024fireredtts}
Hao-Han Guo, Yao Hu, Kun Liu, Fei-Yu Shen, Xu Tang, Yi-Chen Wu, Feng-Long Xie, Kun Xie, and Kai-Tuo Xu.
\newblock Fireredtts: A foundation text-to-speech framework for industry-level generative speech applications.
\newblock \emph{arXiv preprint arXiv:2409.03283}, 2024.

\bibitem[Ho and Salimans(2022)]{ho2022classifier}
Jonathan Ho and Tim Salimans.
\newblock Classifier-free diffusion guidance.
\newblock \emph{arXiv preprint arXiv:2207.12598}, 2022.

\bibitem[Kim et~al.(2025)Kim, Choi, Kim, Jung, and Chung]{kim2025faces}
Ji-Hoon Kim, Jeongsoo Choi, Jaehun Kim, Chaeyoung Jung, and Joon~Son Chung.
\newblock From faces to voices: Learning hierarchical representations for high-quality video-to-speech.
\newblock In \emph{CVPR}, pages 15874--15884, 2025.

\bibitem[Kong et~al.(2024)Kong, Tian, Zhang, Min, Dai, Zhou, Xiong, Li, Wu, Zhang, et~al.]{kong2024hunyuanvideo}
Weijie Kong, Qi Tian, Zijian Zhang, Rox Min, Zuozhuo Dai, Jin Zhou, Jiangfeng Xiong, Xin Li, Bo Wu, Jianwei Zhang, et~al.
\newblock Hunyuanvideo: A systematic framework for large video generative models.
\newblock \emph{arXiv preprint arXiv:2412.03603}, 2024.

\bibitem[Kong et~al.(2025)Kong, Gao, Zhang, Kang, Wei, Cai, Chen, and Luo]{kong2025let}
Zhe Kong, Feng Gao, Yong Zhang, Zhuoliang Kang, Xiaoming Wei, Xunliang Cai, Guanying Chen, and Wenhan Luo.
\newblock Let them talk: Audio-driven multi-person conversational video generation.
\newblock \emph{arXiv preprint arXiv:2505.22647}, 2025.

\bibitem[Lee et~al.(2022)Lee, Ping, Ginsburg, Catanzaro, and Yoon]{lee2022bigvgan}
Sang-gil Lee, Wei Ping, Boris Ginsburg, Bryan Catanzaro, and Sungroh Yoon.
\newblock Bigvgan: A universal neural vocoder with large-scale training.
\newblock \emph{arXiv preprint arXiv:2206.04658}, 2022.

\bibitem[Li et~al.(2025)Li, Xu, Zhan, Mu, Li, Cheng, Chen, Chen, Ye, Wang, et~al.]{li2025openhumanvid}
Hui Li, Mingwang Xu, Yun Zhan, Shan Mu, Jiaye Li, Kaihui Cheng, Yuxuan Chen, Tan Chen, Mao Ye, Jingdong Wang, et~al.
\newblock Openhumanvid: A large-scale high-quality dataset for enhancing human-centric video generation.
\newblock In \emph{CVPR}, pages 7752--7762, 2025.

\bibitem[Liao et~al.(2024)Liao, Wang, Li, Cheng, Zhang, Zhou, and Xing]{liao2024fish}
Shijia Liao, Yuxuan Wang, Tianyu Li, Yifan Cheng, Ruoyi Zhang, Rongzhi Zhou, and Yijin Xing.
\newblock Fish-speech: Leveraging large language models for advanced multilingual text-to-speech synthesis.
\newblock \emph{arXiv preprint arXiv:2411.01156}, 2024.

\bibitem[Lin et~al.(2025)Lin, Jiang, Yang, Zheng, Liang, Zhang, and Liu]{lin2025omnihuman}
Gaojie Lin, Jianwen Jiang, Jiaqi Yang, Zerong Zheng, Chao Liang, Yuan Zhang, and Jingtuo Liu.
\newblock Omnihuman-1: Rethinking the scaling-up of one-stage conditioned human animation models.
\newblock In \emph{CVPR}, pages 13847--13858, 2025.

\bibitem[Lipman et~al.(2022)Lipman, Chen, Ben-Hamu, Nickel, and Le]{lipman2022flow}
Yaron Lipman, Ricky~TQ Chen, Heli Ben-Hamu, Maximilian Nickel, and Matt Le.
\newblock Flow matching for generative modeling.
\newblock \emph{arXiv preprint arXiv:2210.02747}, 2022.

\bibitem[Liu et~al.(2025)Liu, Li, Chen, Wu, Zheng, Ji, Zhou, Jiang, Luo, Fei, et~al.]{liu2025javisdit}
Kai Liu, Wei Li, Lai Chen, Shengqiong Wu, Yanhao Zheng, Jiayi Ji, Fan Zhou, Rongxin Jiang, Jiebo Luo, Hao Fei, et~al.
\newblock Javisdit: Joint audio-video diffusion transformer with hierarchical spatio-temporal prior synchronization.
\newblock \emph{arXiv preprint arXiv:2503.23377}, 2025.

\bibitem[Low et~al.(2025)Low, Wang, and Katyal]{low2025ovi}
Chetwin Low, Weimin Wang, and Calder Katyal.
\newblock Ovi: Twin backbone cross-modal fusion for audio-video generation.
\newblock \emph{arXiv preprint arXiv:2510.01284}, 2025.

\bibitem[Luo et~al.(2023)Luo, Yan, Hu, and Zhao]{luo2023diff}
Simian Luo, Chuanhao Yan, Chenxu Hu, and Hang Zhao.
\newblock Diff-foley: Synchronized video-to-audio synthesis with latent diffusion models.
\newblock \emph{NeurIPS}, 36:\penalty0 48855--48876, 2023.

\bibitem[Peebles and Xie(2023)]{peebles2023scalable}
William Peebles and Saining Xie.
\newblock Scalable diffusion models with transformers.
\newblock In \emph{ICCV}, pages 4195--4205, 2023.

\bibitem[Radford et~al.(2023)Radford, Kim, Xu, Brockman, McLeavey, and Sutskever]{radford2023robust}
Alec Radford, Jong~Wook Kim, Tao Xu, Greg Brockman, Christine McLeavey, and Ilya Sutskever.
\newblock Robust speech recognition via large-scale weak supervision.
\newblock In \emph{International conference on machine learning}, pages 28492--28518. PMLR, 2023.

\bibitem[Rombach et~al.(2022)Rombach, Blattmann, Lorenz, Esser, and Ommer]{rombach2022high}
Robin Rombach, Andreas Blattmann, Dominik Lorenz, Patrick Esser, and Bj{\"o}rn Ommer.
\newblock High-resolution image synthesis with latent diffusion models.
\newblock In \emph{CVPR}, pages 10684--10695, 2022.

\bibitem[Stevens et~al.(1937)Stevens, Volkmann, and Newman]{stevens1937scale}
Stanley~Smith Stevens, John Volkmann, and Edwin~Broomell Newman.
\newblock A scale for the measurement of the psychological magnitude pitch.
\newblock \emph{The journal of the acoustical society of america}, 8:\penalty0 185--190, 1937.

\bibitem[Su et~al.(2024)Su, Ahmed, Lu, Pan, Bo, and Liu]{su2024roformer}
Jianlin Su, Murtadha Ahmed, Yu Lu, Shengfeng Pan, Wen Bo, and Yunfeng Liu.
\newblock Roformer: Enhanced transformer with rotary position embedding.
\newblock \emph{Neurocomputing}, 568:\penalty0 127063, 2024.

\bibitem[Wan et~al.(2025)Wan, Wang, Ai, Wen, Mao, Xie, Chen, Yu, Zhao, Yang, et~al.]{wan2025wan}
Team Wan, Ang Wang, Baole Ai, Bin Wen, Chaojie Mao, Chen-Wei Xie, Di Chen, Feiwu Yu, Haiming Zhao, Jianxiao Yang, et~al.
\newblock Wan: Open and advanced large-scale video generative models.
\newblock \emph{arXiv preprint arXiv:2503.20314}, 2025.

\bibitem[Wang et~al.(2025{\natexlab{a}})Wang, Zuo, Li, Chen, Liao, Zhou, Yin, Dai, Jiang, and Yu]{wang2025universe}
Duomin Wang, Wei Zuo, Aojie Li, Ling-Hao Chen, Xinyao Liao, Deyu Zhou, Zixin Yin, Xili Dai, Daxin Jiang, and Gang Yu.
\newblock Universe-1: Unified audio-video generation via stitching of experts.
\newblock \emph{arXiv preprint arXiv:2509.06155}, 2025{\natexlab{a}}.

\bibitem[Wang et~al.(2025{\natexlab{b}})Wang, Zeng, Qiang, Chen, Wang, Wang, Zhou, Cai, Zhao, Li, et~al.]{wang2025kling}
Jun Wang, Xijuan Zeng, Chunyu Qiang, Ruilong Chen, Shiyao Wang, Le Wang, Wangjing Zhou, Pengfei Cai, Jiahui Zhao, Nan Li, et~al.
\newblock Kling-foley: Multimodal diffusion transformer for high-quality video-to-audio generation.
\newblock \emph{arXiv preprint arXiv:2506.19774}, 2025{\natexlab{b}}.

\bibitem[Wang et~al.(2025{\natexlab{c}})Wang, Wang, Qiang, Deng, Zhang, Zhang, and Gai]{wang2025audiogen}
Le Wang, Jun Wang, Chunyu Qiang, Feng Deng, Chen Zhang, Di Zhang, and Kun Gai.
\newblock Audiogen-omni: A unified multimodal diffusion transformer for video-synchronized audio, speech, and song generation.
\newblock \emph{arXiv preprint arXiv:2508.00733}, 2025{\natexlab{c}}.

\bibitem[Wang et~al.(2025{\natexlab{d}})Wang, Zhang, Qi, Wang, Ji, Xu, Zhang, and Bo]{wang2025omnitalker}
Zhongjian Wang, Peng Zhang, Jinwei Qi, Guangyuan Wang, Chaonan Ji, Sheng Xu, Bang Zhang, and Liefeng Bo.
\newblock Omnitalker: One-shot real-time text-driven talking audio-video generation with multimodal style mimicking.
\newblock \emph{arXiv preprint arXiv:2504.02433}, 2025{\natexlab{d}}.

\bibitem[Wu et~al.(2025)Wu, Zhu, Liu, Zhao, Zhai, Cao, and Zha]{wu2025improved}
Pingyu Wu, Kai Zhu, Yu Liu, Liming Zhao, Wei Zhai, Yang Cao, and Zheng-Jun Zha.
\newblock Improved video vae for latent video diffusion model.
\newblock In \emph{CVPR}, pages 18124--18133, 2025.

\bibitem[Xu et~al.(2025)Xu, Guo, Hu, Chu, Wang, He, Wang, Shi, He, Zhu, et~al.]{xu2025qwen3omni}
Jin Xu, Zhifang Guo, Hangrui Hu, Yunfei Chu, Xiong Wang, Jinzheng He, Yuxuan Wang, Xian Shi, Ting He, Xinfa Zhu, et~al.
\newblock Qwen3-omni technical report.
\newblock \emph{arXiv preprint arXiv:2509.17765}, 2025.

\bibitem[Yang et~al.(2025)Yang, Li, Yang, Zhang, Hui, Zheng, Yu, Gao, Huang, Lv, et~al.]{yang2025qwen3}
An Yang, Anfeng Li, Baosong Yang, Beichen Zhang, Binyuan Hui, Bo Zheng, Bowen Yu, Chang Gao, Chengen Huang, Chenxu Lv, et~al.
\newblock Qwen3 technical report.
\newblock \emph{arXiv preprint arXiv:2505.09388}, 2025.

\bibitem[Yang et~al.(2024)Yang, Teng, Zheng, Ding, Huang, Xu, Yang, Hong, Zhang, Feng, et~al.]{yang2024cogvideox}
Zhuoyi Yang, Jiayan Teng, Wendi Zheng, Ming Ding, Shiyu Huang, Jiazheng Xu, Yuanming Yang, Wenyi Hong, Xiaohan Zhang, Guanyu Feng, et~al.
\newblock Cogvideox: Text-to-video diffusion models with an expert transformer.
\newblock \emph{arXiv preprint arXiv:2408.06072}, 2024.

\bibitem[Zhang et~al.(2025{\natexlab{a}})Zhang, Guo, Yang, Yu, Zhang, Lei, Mai, Yan, Yang, Yang, et~al.]{zhang2025minimax}
Bowen Zhang, Congchao Guo, Geng Yang, Hang Yu, Haozhe Zhang, Heidi Lei, Jialong Mai, Junjie Yan, Kaiyue Yang, Mingqi Yang, et~al.
\newblock Minimax-speech: Intrinsic zero-shot text-to-speech with a learnable speaker encoder.
\newblock \emph{arXiv preprint arXiv:2505.07916}, 2025{\natexlab{a}}.

\bibitem[Zhang et~al.(2025{\natexlab{b}})Zhang, Liu, Zheng, Chen, Ding, and Di]{zhang2025deepaudio}
Haomin Zhang, Chang Liu, Junjie Zheng, Zihao Chen, Chaofan Ding, and Xinhan Di.
\newblock Deepaudio-v1: Towards multi-modal multi-stage end-to-end video to speech and audio generation.
\newblock \emph{arXiv preprint arXiv:2503.22265}, 2025{\natexlab{b}}.

\bibitem[Zhang et~al.(2024)Zhang, Gu, Zeng, Xing, Wang, Wu, and Chen]{zhang2024foleycrafter}
Yiming Zhang, Yicheng Gu, Yanhong Zeng, Zhening Xing, Yuancheng Wang, Zhizheng Wu, and Kai Chen.
\newblock Foleycrafter: Bring silent videos to life with lifelike and synchronized sounds.
\newblock \emph{arXiv preprint arXiv:2407.01494}, 2024.

\bibitem[Zhao et~al.(2023{\natexlab{a}})Zhao, Bai, Rao, Zhou, and Lu]{zhao2023unipc}
Wenliang Zhao, Lujia Bai, Yongming Rao, Jie Zhou, and Jiwen Lu.
\newblock Unipc: A unified predictor-corrector framework for fast sampling of diffusion models.
\newblock \emph{NeurIPS}, 36:\penalty0 49842--49869, 2023{\natexlab{a}}.

\bibitem[Zhao et~al.(2023{\natexlab{b}})Zhao, Gu, Varma, Luo, Huang, Xu, Wright, Shojanazeri, Ott, Shleifer, et~al.]{zhao2023pytorch}
Yanli Zhao, Andrew Gu, Rohan Varma, Liang Luo, Chien-Chin Huang, Min Xu, Less Wright, Hamid Shojanazeri, Myle Ott, Sam Shleifer, et~al.
\newblock Pytorch fsdp: experiences on scaling fully sharded data parallel.
\newblock \emph{arXiv preprint arXiv:2304.11277}, 2023{\natexlab{b}}.

\bibitem[Zhou et~al.(2025)Zhou, Zhou, He, Zhou, Wang, Deng, and Shu]{zhou2025indextts2}
Siyi Zhou, Yiquan Zhou, Yi He, Xun Zhou, Jinchao Wang, Wei Deng, and Jingchen Shu.
\newblock Indextts2: A breakthrough in emotionally expressive and duration-controlled auto-regressive zero-shot text-to-speech.
\newblock \emph{arXiv preprint arXiv:2506.21619}, 2025.

\end{thebibliography}
